\documentclass[runningheads]{llncs}
\usepackage{graphicx}
\usepackage{comment}
\usepackage{amsmath,amssymb} % define this before the line numbering.
\usepackage{color}

\usepackage{multirow}

\usepackage{bbold}

\begin{document}
% \renewcommand\thelinenumber{\color[rgb]{0.2,0.5,0.8}\normalfont\sffamily\scriptsize\arabic{linenumber}\color[rgb]{0,0,0}}
% \renewcommand\makeLineNumber {\hss\thelinenumber\ \hspace{6mm} \rlap{\hskip\textwidth\ \hspace{6.5mm}\thelinenumber}}
% \linenumbers
\pagestyle{headings}
\mainmatter
\def\ECCVSubNumber{4505}  % Insert your submission number here

\title{Zero-Shot Recognition through Image-Guided Semantic Classification} % Replace with your title

% INITIAL SUBMISSION 
\begin{comment}
\titlerunning{ECCV-20 submission ID \ECCVSubNumber} 
\authorrunning{ECCV-20 submission ID \ECCVSubNumber} 
\author{Anonymous ECCV submission}
\institute{Paper ID \ECCVSubNumber}
\end{comment}
%******************

% CAMERA READY SUBMISSION
%\begin{comment}
\titlerunning{Zero-Shot Recognition through Image-Guided Semantic Classification}
% If the paper title is too long for the running head, you can set
% an abbreviated paper title here
%
\author{Mei-Chen Yeh\orcidID{0000-0001-8665-7860} \and
Fang Li}
\authorrunning{M. Yeh and F. Li}
% First names are abbreviated in the running head.
% If there are more than two authors, 'et al.' is used.
%
\institute{National Taiwan Normal University, Taipei, Taiwan \\
\email{myeh@csie.ntnu.edu.tw}\\
}
%\end{comment}
%******************
\maketitle

%%%%%%%%% ABSTRACT
\begin{abstract}
We present a new embedding-based framework for zero-shot learning (ZSL). Most embedding-based methods aim to learn the correspondence between an image classifier (visual representation) and its class prototype (semantic representation) for each class. Motivated by the binary relevance method for multi-label classification, we propose to inversely learn the mapping between an image and a semantic classifier. Given an input image, the proposed Image-Guided Semantic Classification (IGSC) method creates a label classifier, being applied to all label embeddings to determine whether a label belongs to the input image. Therefore, semantic classifiers are image-adaptive and are generated during inference. IGSC is conceptually simple and can be realized by a slight enhancement of an existing deep architecture for classification; yet it is effective and outperforms state-of-the-art embedding-based generalized ZSL approaches on standard benchmarks. %Code will be made publicly available upon publication.

\end{abstract}

%%%%%%%%% BODY TEXT
\section{Introduction}

\begin{figure*}[t]
\begin{center}
 \includegraphics[width=0.85\linewidth]{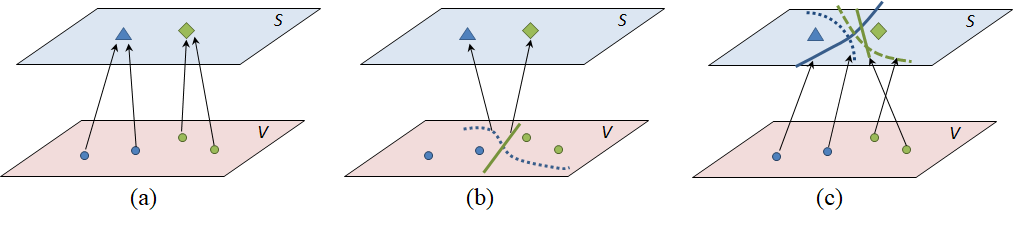}
\end{center}
   \caption{Zero-shot learning paradigms. (a) Conventional visual-to-semantic mapping trained on classification loss. (b) Another interpretation of visual-to-semantic mapping between visual and semantic representations. (c) The proposed IGSC, aiming to learn the correspondence between an image and a semantic classifier.}
\label{fig:pagadigm}
\end{figure*}

As a feasible solution for addressing the limitations of supervised classification methods, zero-shot learning (ZSL) aims to recognize objects whose instances have not been seen during training \cite{Larochelle08,Palatucci2009ZeroshotLW}. Unseen classes are recognized by associating seen and unseen classes through some form of \emph{semantic space}; therefore, the knowledge learned from seen classes is transferred to unseen classes. In the semantic space, each class has a corresponding vector representation called a \emph{class prototype}. Class prototypes can be obtained from human-annotated attributes that describe visual properties of objects \cite{Farhadi2009DescribingOB,AWA} or from word embeddings learned in an unsupervised manner from text corpus \cite{word2vec,Glove,devlin2018bert}.

A majority of ZSL methods can be viewed using the visual-semantic embedding framework, as displayed in Figure~\ref{fig:pagadigm} (a). Images are mapped from the visual space to the semantic space in which all classes reside. Then, the inference is performed in this common space. For example, an image is assigned to the nearest class prototype in the semantic space \cite{Akata2013LabelEmbeddingFA,Frome2013DeViSEAD,Socher2013ZeroShotLT}. Although class embedding has rich semantic meanings, each class is represented by only a single class prototype to determine where images of that class collapse inevitably \cite{MarcoBaroni16,Fu2015TransductiveMZ}. According to the hubness phenomenon, the mapped semantic representations from images collapse to hubs, which are close to many other points in the semantic space, rather than being similar to the true class label \cite{MarcoBaroni16}.

%Another perspective of embedding-based ZSL methods is to construct a classifier for unseen classes by using the correspondence between a binary one-versus-rest image classifier for each class (i.e., visual representation of a class) and its class prototype in the semantic space (i.e., semantic representation of a class) \cite{Wang2019ASO}. For example, a commonly used choice for such correspondence is the bilinear function $\theta(x_i)^TW\phi(y_j)$ \cite{Frome2013DeViSEAD,Akata2013LabelEmbeddingFA,Akata2015EvaluationOO,RomeraParedes2015AnES,DSSC}, where $\theta(x_i)$ and $\phi(y_j)$ denote the visual and semantic embeddings, respectively, and $W$ denotes model parameters. For class $y_j$, the parameter of the binary one-versus-rest image classifier is $W\phi(y_j)$ (a linear classifier). Considerable efforts have been made to extend the linear function to nonlinear ones \cite{Xian2016LatentEF,Wang18,Elhoseiny17,Qiao16}. Figure~\ref{fig:pagadigm}(b) illustrates this view.
Another perspective of embedding-based ZSL methods is to construct an image classifier for each unseen class by learning the correspondence between a binary one-versus-rest image classifier for each class (i.e., visual representation of a class) and its class prototype in the semantic space (i.e., semantic representation of a class) \cite{Wang2019ASO}. Once this correspondence function is learned, a binary one-versus-rest image classifier can be constructed for an unseen class with its prototype \cite{Wang2019ASO}. For example, a commonly used choice for such correspondence is the bilinear function \cite{Frome2013DeViSEAD,Akata2013LabelEmbeddingFA,Akata2015EvaluationOO,RomeraParedes2015AnES,DSSC}. Considerable efforts have been made to extend the linear function to nonlinear ones \cite{Xian2016LatentEF,Wang18,Elhoseiny17,Qiao16}. Figure~\ref{fig:pagadigm} (b) illustrates this perspective.

%Learning the correspondence between an image classifier and a class prototype involves the following drawbacks. First, the assumption of using a single image classifier for each class is unrealistic because there are several manners for separating classes in both visual and semantic spaces. We argue that semantic classification should be conducted dynamically conditioned on an input image. For example, the visual attribute \textit{wheel} may be useful for classifying most car images. Nevertheless, cars with missing wheels should also be correctly recognized using other visual attributes. Second, the number of training pairs for learning the correspondence is constrained to be the number of class labels, which hinders the robustness of deep learning models that usually require large-scale training data. 
Learning the correspondence between an image classifier and a class prototype involves the following drawbacks. First, the assumption of using a single image classifier for each class is unrealistic because the manner for separating classes in both visual and semantic spaces would not be unique. We argue that semantic classification should be conducted dynamically conditioned on an input image. For example, the visual attribute \textit{wheel} may be useful for classifying most car images. Nevertheless, cars with missing wheels should also be correctly recognized using other visual attributes. Therefore, image-specific semantic classifiers make better sense than category-specific ones because the classifier weights can be adaptively determined based on image content. Second, the number of training pairs for learning the correspondence is constrained to be the number of class labels. In other words, a training set with $C$ labels has only $C$ visual-semantic classifier pairs to build the correspondence. This may hinder the robustness of deep learning models that usually require large-scale training data. 

In this paper, we present an Image-Guided Semantic Classification (IGSC) method to address these problems. This method aims to inversely learn the correspondence between an image and its corresponding label classifier, as illustrated in Figure~\ref{fig:pagadigm} (c). In contrast to methods in previous studies \cite{Fast0Tag,Frome2013DeViSEAD,Socher2013ZeroShotLT}, the IGSC method is \emph{not} used to map an image from the visual space to the semantic space. Instead, this method learns from an image and seeks for combinations of variables in the semantic space (e.g., combinations of attributes) that distinguish a class from other classes. As will be demonstrated later in this paper, the correspondence between an image and a semantic classifier learned from seen classes can be effectively transferred to recognize unseen concepts. Compared with state-of-the-art ZSL methods, the IGSC method is conceptually simple and can be implemented using a simple network architecture. In addition, it is more powerful than many existing deep learning based models for generalized zero-shot recognition. The proposed IGSC method has the following characteristics:
\begin{itemize}
%\item We present a new model to learn the correspondence between an image in the visual space and a classifier in the semantic space. Semantic classifiers are generated dynamically based on input images during testing. This model has increased flexibility because of its adaptive nature. The correspondence can be learned with training pairs in the scale of training images rather than that of classes.
\item The IGSC method learns the correspondence between an image in the visual space and a classifier in the semantic space. The correspondence can be learned with training pairs in the scale of training images rather than that of classes.
\item The IGSC method performs learning to learn in an end-to-end manner. Label classification is conducted by an image-guided semantic classifier whose weights are generated based on the input image. This model is simple yet powerful because of its adaptive nature.
\item The IGSC method unifies visual attribute detection and label classification. This is achieved via a conditional network (the proposed classifier learning method), in which label classification is the main task of interest and the conditional input image provides additional information of a specific situation.
\item The IGSC method is flexible in that it can be developed using state-of-the-art network structures. To the best of our knowledge, the IGSC model is the first ZSL model that learns model representations. We hope that it will bring a different perspective to the ZSL problem, gaining in a deeper understanding of knowledge transfer. %We validated the effectiveness of the proposed method on four popular benchmarks and achieved state-of-the-art performance levels. 
\end{itemize}

We evaluated the proposed method with experiments conducted on the public ZSL benchmark datasets, including SUN \cite{Patterson2012SUNAD}, CUB \cite{Patterson2012SUNAD}, AWA2 \cite{AWA}, and aPY \cite{Farhadi2009DescribingOB}. Experimental results demonstrated that the proposed method achieved promising performance, compared with current state-of-the-art methods. We explain and empirically analyze the superior performance of the method in the Discussion section. The remainder of the paper is organized as follows: We briefly review work related to zero-shot recognition in Section \ref{sec:related}. Section \ref{sec:method} presents the details of the proposed framework. The experimental results and conclusions are provided in Sections \ref{sec:exp} and \ref{sec:conclusion}, respectively.

\section{Related Work}
\label{sec:related}

Zero-shot learning has evolved rapidly during the last decade, and therefore documenting the extensive literature with limited pages is rarely possible. In this section, we review a few representative zero-shot learning methods and refer the interested readers to~\cite{Xian19TPAMI,Wang2019ASO} for a comprehensive survey. One pioneering main stream of ZSL uses attributes to infer the label of an image belonging to one of the unseen classes \cite{AWA,Halah16,Norouzi2013ZeroShotLB,Jayaraman14,Kankuekul12}. The attributes of an image are predicted, then the class label is inferred by searching the class which attains the most similar set of attributes. For example, the Direct Attribute Prediction (DAP) model \cite{Lampert2009LearningTD} first estimates the posterior of each attribute for an image by learning probabilistic attribute classifiers. A test sample is then classified by each attribute classifier alternately, and the class label is predicted by probabilistic estimation. Similar to the attribute-based methods, the proposed IGSC method has the merits of modeling the relationships among classes. However, to the best of our knowledge, IGSC is the first ZSL method that unifies these two steps: attribute classifier learning and inferring from detected attributes to the class. Furthermore, attribute classifiers are jointly learned in IGSC.

A broad family of ZSL methods apply an embedding framework that directly learns a mapping from the visual space to the semantic space \cite{Palatucci2009ZeroshotLW,Akata2013LabelEmbeddingFA,Akata2015EvaluationOO,RomeraParedes2015AnES}. The visual-to-semantic mapping can be linear \cite{Frome2013DeViSEAD} or nonlinear \cite{Socher2013ZeroShotLT}. For example, DeViSE \cite{Frome2013DeViSEAD} learns a linear mapping between the image and semantic spaces using an efficient ranking loss formulation. CMT \cite{Socher2013ZeroShotLT} uses a neural network with two hidden layers to learn a nonlinear projection from image feature space to word vector space. More recently, deep neural network models are proposed to mirror learned semantic relations among classes in the visual domain from the image~\cite{Annadani2018PreservingSR} or from the part~\cite{Zhu2018GeneralizedZR} levels. The proposed IGSC model is also an embedding-based ZSL method that builds the correspondence between a visual and a semantic space. IGSC differs significantly from existing methods in that IGSC learns the correspondence between an image and its semantic classifier, enabling the possibility of using different classification manners to separate class prototypes in the semantic space. Even though each class has only one class prototype, the classification is \emph{not} performed by nearest neighbor search and therefore suffers much less from the hubness problem.

Recent ZSL models adopt the generative adversarial network (GAN)~\cite{GAN} or other generative models for synthesizing unseen examples \cite{Bucher17,Long17,Jiang18,Verma18,Xian18,Zhu18} or for reconstructing training images \cite{Chen18}. The synthesized images obtained at the training stage can be fed to conventional classifiers so that ZSL is converted into the conventional supervised learning problem \cite{Long17}. The transformation from attributes to image features require involving generative models such as denoising autoencoders \cite{Bucher17}, GAN \cite{Xian18,Zhu18} or their variants \cite{Verma18}. Despite outstanding performances reported in the papers, these works leverage some form of the unseen class information during training. In view of real-world applications involving recognition in-the-wild, novel classes including the image samples as well as the semantic representations may not be available in model learning. The proposed method is agnostic to all unseen class information during training. Furthermore, the proposed method is much simpler in the architecture design and has a much smaller model size, compared with the generative methods.

It is worth noting that the idea of predicting classifiers has been used in \cite{Wang2018ZeroShotRV}. Given a learned knowledge graph, Wang \textit{et al.} take as input semantic embeddings for each node (representing visual category) and predict a visual classifier for each category through a series of graph convolutions. While Wang \textit{et al.} predict the visual classifier for each category \cite{Wang2018ZeroShotRV}, this paper differs fundamentally in that we predict the semantic classifier for each image.

\section{Approach}
\label{sec:method}

We start with formulating the ZSL problem and describe in detail the model design and training. 

\subsection{Problem Description}
Given a training set $S = \{(x_n, y_n), n = 1 \dots N \}$, with $y_n \in \mathcal{Y}_s$ being a class label in the seen class set, the goal of ZSL is to learn a classifier $f: \mathcal{X} \rightarrow \mathcal{Y}$ which can generalize to predict any image $x$ to its correct label, which is not only in $\mathcal{Y}_s$ but also in the unseen class set $\mathcal{Y}_u$. In the prevailing family of compatibility learning ZSL \cite{Xian19TPAMI,Ba15}, the prediction is made via:
\begin{equation}
\label{equ:prediction}
	\hat{y} = f(x; W) = \mathop{\arg\max}_{y \in \mathcal{Y}} \ F(x, y; W). 
\end{equation}
In particular, if $\mathcal{Y} \in \mathcal{Y}_u$, this is the conventional ZSL setting; if $\mathcal{Y} \in \mathcal{Y}_s \cup \mathcal{Y}_u$, this is the generalized zero-shot learning (GZSL) setting, which is more practical for real-world applications. The compatibility function $F(\cdot)$---parameterized by $W$---is used to associate visual and semantic information.

In the visual space, each image $x$ has a vector representation, denoted by $\theta(x)$. Similarly, each class label $y$ has a vector representation in the semantic space (called the class prototype), denoted by $\phi(y)$. In short, $\theta(x)$ and $\phi(y)$ are the image and class embeddings, both of which are given.

\begin{figure*}[t]
\begin{center}
 \includegraphics[width=0.9\linewidth]{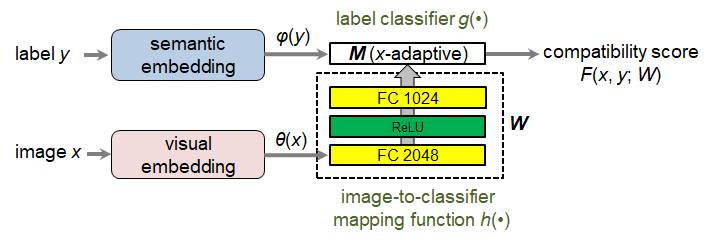}
\end{center}
   \caption{The architecture of IGSC. This model receives an image and a label, and it returns the compatibility score of this input pair. The score indicates the probability of the label belonging to the image. The score is calculated by a label classifier $g(\cdot)$, whose weights $M$ are stored in the output layer of a fully connected neural network. Therefore, weight values depend on the input image. The neural network is characterized by the parameters $W$, which are the only parameters required to learn from training data.}
\label{fig:model}
\end{figure*}

\subsection{Image-Guided Semantic Classification Model}
The compatibility function in this work is achieved by implementing two functions, $h(\theta(x); W)$ and $g(\phi(y); M)$, as illustrated in Figure~\ref{fig:model}. The first function $h(\cdot)$ receives an image embedding as input and returns parameters $M$ characterizing a label classifier:
\begin{equation}
	M = h(\theta(x); W).
\end{equation}
In other words, $h(\cdot)$ learns the mapping between image representations and their associate semantic classifiers. Each image has its own semantic classifier. Images of the same class may have different classifier weights. 

Different from existing methods where the classifier weights are part of model parameters and thereby being static after training, the classifier weights in IGSC are dynamically generated during test time. The semantic classifiers are created conditioned on the input image.

The second function $g(\cdot)$ is a label classifier, characterized by the parameters outputted by $h(\cdot)$. This function takes a label vector as input, and returns a prediction score indicating the probability of the label belonging to the input image:
\begin{equation}
	s = g(\phi(y); M).
\end{equation}
Let $s_j$ denote the prediction score for a label $j$. In multi-class (single-label) image classification, the final compatibility score is obtained by normalizing the prediction scores to probabilistic values with softmax:
\begin{equation}
\label{equ:softmax}
   %p(\hat{y} = j \mid x) = \frac{\exp(s_j)}{\sum_{k=1}^{\vert \mathcal{Y}_s\vert} \exp(s_k)}.
   F(x, y_j; W) = \frac{\exp(s_j)}{\sum_{k=1}^{\vert \mathcal{Y}\vert} \exp(s_k)}.
\end{equation}
The test image is assigned to the class with the highest compatibility score. In multi-label image classification, we replace softmax by a sigmoid activation function. The prediction is made by choosing labels whose compatibility score is greater than a threshold.

For clarity, we make a distinction between the parameters of these two functions: \emph{model parameters} ($W$) and \emph{dynamically generated parameters} ($M$). Model parameters $W$ denote the layer parameters (i.e., two fully connected layers in our implementation) initialized and updated during training. These parameters are static during test time for all samples. Dynamically generated parameters $M$ are produced on-the-fly and input-specific. $M$ are dynamically generated during test time and characterize a label classifier $g(\cdot)$. We emphasize again that $W$ are the only model parameters required to learn from training data.

It is worth noting that the mechanism of image-guided semantic classification is similar to that of \textit{Dynamic Filter Networks}~\cite{Dynamic_filter_network}, in which the filters are generated dynamically conditioned on an input. A similar mechanism also appears in~\cite{dynamic_conditional_networks}, which predicts a set of adaptive weights from conditional inputs to linearly combine the basis filters. The proposed method differs fundamentally in that both~\cite{Dynamic_filter_network} and~\cite{dynamic_conditional_networks} focus on learning image representations, while our method aims to learn model representations that are applied to a different modality (i.e., labels).

\subsection{Forms of Label Classifiers}
The image-guided label classifier can be either linear or nonlinear, which receives a label embedding and returns a prediction score of the label. The label classifier is obtained by feeding an image into a visual model (e.g., AlexNet \cite{AlexNet}, VGGNet \cite{VGGNet}, or other deep neural networks), followed by two fully connected layers and an output layer. The dimension of the output layer is set to accommodate the label classifier weights. 

In this study we experiment with two variations of the label classifier. The linear label classifier is represented as:
\begin{equation}
\label{equ:linear}
	g(\phi(y); M) = \mathbf{m}\phi(y) + b.
\end{equation}
where $\mathbf{m} \in \mathbb{R}^d$ is a weight vector, $b$ is a threshold and $M = (\mathbf{m}, b)$. The dimension $d$ is set to that of the label vector (e.g., $d$ = 300 if using 300-dim word2vec~\cite{word2vec}). Alternatively, the nonlinear label classifier is implemented using a two-layer neural network:
\begin{equation}
\label{equ:nonlinear}
	g(\phi(y); M) = \mathbf{m_2} \tanh(\mathbf{M_1}\phi(y) + b_1) + b_2,
\end{equation}
where $\mathbf{M_1} \in  \mathbb{R}^{h \times d}, \mathbf{m_2} \in  \mathbb{R}^h$ and $M = (\mathbf{M_1}, b_1, \mathbf{m_2}, b_2)$. The nonlinear classifier characterizes the $d$-dim semantic space by using $h$ perceptrons and performs the classification task. As will be shown in Section \ref{sec:exp}, the nonlinear label classifier outperforms a linear one. We would like to highlight again that the label classifier weights $M$ are created during inference. This image-dependent label classifier seeks a good combination of variables in the semantic space  for distinguishing ground-truth class from other classes.

For GZSL, it is beneficial to enable \emph{calibrated stacking} \cite{Chao2016AnES}, which reduces the scores for seen classes. This leads to the following modification: 
\begin{equation}
\label{equ:calibration}
	\hat{y} = \mathop{\arg\max}_{y \in \mathcal{Y}_s \cup \mathcal{Y}_u} \ \big( g(\phi(y); M) - \gamma \mathbb{1}{[y \in \mathcal{Y}_s]} \big), 
\end{equation}
where $\mathbb{1}{[y \in \mathcal{Y}_s]} \in \{0, 1\}$ indicates whether or not $y$ is a seen class and $\gamma$ is a calibration factor.

\subsection{Learning Model Parameters}
Recall that the objective of ZSL is to correctly assign a test image to its label. This is a typcial classification problem. For a training sample $x_i$, Let $y_i = \{y_i^1, y_i^2, ..., y_i^{\vert \mathcal{Y}_s \vert} \} \in \{0, 1\}$ denote the one-hot encoding of the ground truth label and $p_i = \{p_i^1, p_i^2, ..., p_i^{\vert \mathcal{Y}_s \vert} \}$  denote the compatibility scores of $x_i$ (Equ.~\ref{equ:softmax}). That is, $p_i^j = F(x_i, y_j; W)$. The model parameters $W$ are learned by minimizing the cross entropy loss: 
\begin{equation}
\mathcal L = - \sum_{i=1}^{N}\sum_{j=1}^{\vert \mathcal{Y}_s \vert} y_i^j \log (p_i^j) + (1-y_i^j) \log (1-p_i^j).
\end{equation}
The weights including $W$ and those of the image/semantic embedding networks can be jointly learned end-to-end; however, the results reported in Section \ref{sec:exp} were obtained by freezing the weights of feature extractors for a fair comparison. That is, all methods under comparison used the same image and semantic representations in the experiments.

\subsection{Training Details}

We used Adaptive Moment Estimation (Adam) for optimizing the model. We augmented the data by random cropping and mirroring. The learning rate was set fixed to $10^{-5}$. Training time for a single epoch ranged from 91 seconds to 595 seconds (depending on which dataset was used). Training the models using four benchmark datasets roughly took 11 hours in total. The runtime was reported running on a machine with an Intel Core i7-7700 3.6-GHz CPU, NVIDIA's GeForce GTX 1080Ti and 32 GB of RAM. The dimension $h$ in the nonlinear variant of the semantic classifier $g(\cdot)$ was set to 30 in the experiments.

\section{Experiments}
\label{sec:exp}

This section presents the experimental results. We compare the proposed approach with state-of-the-art methods using four benchmarks, including SUN \cite{Patterson2012SUNAD}, CUB \cite{Patterson2012SUNAD}, AWA2 \cite{AWA}, and aPY \cite{Farhadi2009DescribingOB}. Please note that all methods under comparison---including the proposed method---use the \textit{class-inductive instance-inductive} (CIII) setting~\cite{Wang2019ASO}. Only labeled training instances and class prototypes of seen classes are available. This is the most restricted setting. Alternatively, methods that are transductive for unseen class prototypes (class-transductive instance-inductive, or CTII) and unlabeled unseen test instances (class-transductive instance-transductive, or CTIT), can achieve better performances because more information is involved in model learning. For example, recent generative models in the inductive setting are only inductive to samples~\cite{Xian2019fVAEGAND2AF}. They are CTII methods. These methods use unseen class labels during training, which is different to our setting and, therefore, are not compared.

\subsection{Datasets and Experimental Setting}

We used four benchmark datasets described below and summarized in Table~\ref{TABLE1}. We followed the new split provided by \cite{Xian19TPAMI} because this split ensured that classes at test should be strictly unseen at training.

\noindent
\textbf{SUN Attribute (SUN)} \cite{Patterson2012SUNAD} is a fine-grained scene dataset, containing 14,340 images from 717 types of scenes annotated with 102 attributes. The train split has 10,320 images from 645 classes (65 classes for validation). The test split has 2,580 images from the 645 seen classes and 1,440 images from the 72 unseen classes.

\noindent
\textbf{Caltech-UCSD-Birds-200-2011 (CUB)} \cite{Patterson2012SUNAD} is a fine-grained dataset, containing 11,788 images from 200 different types of birds annotated with 312 attributes. The train split has 7,057 images across 150 classes (50 classes for validation). The test split has 1,764 images from the 150 seen classes and 2,967 images from the 50 unseen classes.

\noindent
\textbf{Animals with Attributes (AWA)} \cite{AWA} is a coarse-grained dataset, containing 37,322 images from 50 animal classes with at least 92 labeled examples per class. We used the AWA2 released by \cite{Xian19TPAMI} as the images from the original AWA are restricted due to photo copyright reasons. The train split has 23, 527 images from 40 classes (13 classes for validation). The test split has 5,882 images from the 40 seen classes and 7,913 images from the 10 unseen classes.

\noindent
\textbf{Attribute Pascal and Yahoo (aPY)} \cite{Farhadi2009DescribingOB} is a coarse-grained dataset, containing 15,339 images from 32 classes annotated with 64 attributes. The train split has 5,932 images from 20 classes (5 classes for validation). The test split has 1,483 images from the 20 seen classes and 7,924 from the 12 unseen classes.

\paragraph{Visual and semantic embeddings.}

For a fair comparison, we used the 2048-dimensional ResNet-101 features provided by \cite{Xian19TPAMI} as image representations. For label representations, we used the semantic embeddings provided by \cite{Xian19TPAMI}, each of which is an L2-normalized attribute vector. Note that the proposed method can use other methods as visual (e.g., ResNet-152) or class (e.g., word2vec \cite{word2vec}, BERT \cite{devlin2018bert}) embeddings.

\paragraph{Evaluation protocols.}

We followed the standard evaluation metrics used in the literature. For ZSL, we used average per-class top-1 accuracy as the evaluation metric, where the prediction (Eq.~\ref{equ:prediction}) is successful if the predicted class is the correct ground truth. For GZSL, we reported $acc_s$ (test images are from seen classes and the prediction labels are the union of seen and unseen classes) and $acc_u$ (test images are from unseen classes and the prediction labels are the union of seen and unseen classes). We computed the harmonic mean \cite{Xian19TPAMI} of accuracy rates on seen classes $acc_s$ and unseen classes $acc_u$:
\begin{equation}
   H = \frac{2 \times acc_s \times acc_u}{acc_s + acc_u}.
\end{equation}
The harmonic mean offers a comprehensive metric in evaluating GZSL methods. The harmonic mean value is high only when both accuracy rates are high.

For a fair comparison, we reported the average results of three random trials for each ZSL and GZSL experiment. 

% table 1
\begin{table*}[ht]
\centering
\caption{Summary of the datasets used in the experiments.}\smallskip
\label{TABLE1}
\resizebox{0.95\textwidth}{!}{
\begin{tabular}{*{2}{c|}*{2}{c}|*{4}{c}}
 & & \multicolumn{2}{c|}{Number of classes} &
\multicolumn{4}{c}{Number of samples} \\
Dataset & Embedding dim. & Seen & Unseen & Training & Test (seen) & Test (unseen) & Total\\
\hline
SUN \cite{Patterson2012SUNAD} & 102 & 580 + 65 & 72 & 10,320 & 2,580 & 1,440 & 14,340\\
CUB \cite{Welinder2010CaltechUCSDB2} & 312 & 100 + 50 & 50 & 7,057 & 2,967 & 1,764 & 11,788\\
AWA2 \cite{AWA} & 85 & 27 + 13 & 10 & 23,527 & 5,882 & 7,913 & 37,322\\
aPY \cite{Farhadi2009DescribingOB} & 64 & 15 + 5 & 12 & 5,932 & 1,483 & 7,924 & 15,339\\
\end{tabular}
}
\end{table*}

% table 2-1
\begin{table}[ht]
\centering
\caption{Standard zero-shot learning results (top-1 accuracy) on four benchmark datasets. Results of the existing approaches are taken from~\cite{Xian19TPAMI}.}\smallskip
\label{tab:zsl}
\begin{tabular}{l|*{4}{c}}
% & {SUN} & {CUB} & {c}{AWA2} & {aPY}\\
 & {SUN} & {CUB} & {AWA2} & {aPY}\\
% \cline{2-13}
Method & $acc$ & $acc$ & $acc$ & $acc$\\
\hline
DAP\cite{Lampert2009LearningTD} & 39.9 & 40.0 & 46.1 & 33.8\\
IAP\cite{Lampert2009LearningTD} & 19.4 & 24.0 & 35.9 & 36.6\\
CONSE\cite{Norouzi2013ZeroShotLB} & 38.8 & 34.3 & 44.5 & 26.9\\
CMT\cite{Socher2013ZeroShotLT} & 39.9 & 34.6 & 37.9 & 28.0\\
SSE\cite{Zhang2015ZeroShotLV} & 51.5 & 43.9 & 61.0 & 34.0\\
LATEM\cite{Xian2016LatentEF} & 55.3 & 49.3 & 55.8 & 35.2\\
ALE\cite{Akata2013LabelEmbeddingFA} & 58.1 & 54.9 & 62.5 & 39.7\\
DEVISE\cite{Frome2013DeViSEAD} & 56.5 & 52.0 & 59.7 & \textbf{39.8}\\
SJE\cite{Akata2015EvaluationOO} & 53.7 & 53.9 & 61.9 & 32.9\\
ESZSL\cite{RomeraParedes2015AnES} & 54.5 & 53.9 & 58.6 & 38.3\\
SYNC\cite{Changpinyo2016SynthesizedCF} & 56.3 & 55.6 & 46.6 & 23.9\\
SAE\cite{Kodirov2017SemanticAF} & 40.3 & 33.3 & 54.1 & 8.3\\
GFZSL\cite{Verma2017ASE} & \textbf{60.6} & 49.3 & \textbf{63.8} & 38.4\\
%SP-AEN \cite{Chen18} & 59.2 & 55.4 & 58.5 & 24.1\\
%PSRZSL\cite{Annadani2018PreservingSR} & 61.4 & 56.0 & 63.8 & 38.4\\
%AREN \cite{Xie2019AttentiveRE} & 60.6 & 71.8 & 67.9 & 39.2\\
%GZSRVSE \cite{Zhu2018GeneralizedZR} & - & 66.7 & 69.1 & 50.1\\
%f-VAEGAN-D2 \cite{Xian2019fVAEGAND2AF} & 65.6 & 72.9 & 70.3 & -\\
\hline
IGSC (linear) & 55.4 & 51.9 & 58.2 & 36.5\\
%Ours-linear(Res152) & 56.7 & 53.0 & 50.9 & 37.8\\
IGSC (nonlinear) & 58.3 & \textbf{56.9} & 62.1 & 35.2\\
%Ours-nonlinear(Res152) & 59.8 & \textbf{57.3} & 63.3 & 33.8 & \textbf{23.7} & 36.1 & \textbf{28.6} & \textbf{28.6} & \textbf{68.0} & \textbf{40.3} & \textbf{22.7} & 83.9 & \textbf{35.7} & \textbf{14.9} & 67.6 & \textbf{24.4}\\
\end{tabular}
\end{table}

% table 2-2
\begin{table*}[ht]
\centering
\caption{Generalized zero-shot learning results (top-1 accuracy and H) on four benchmark datasets. All methods are agnostic to both unseen images and unseen semantic vectors during training.}\smallskip
\label{tab:gzsl}
% \resizebox{0.95\textwidth}{!}{
\begin{tabular}{l|*{12}{r}}
 & \multicolumn{3}{c}{SUN} & \multicolumn{3}{c}{CUB} & \multicolumn{3}{c}{AWA2} & \multicolumn{3}{c}{aPY}\\
% \cline{2-13}
Method &$acc_u$ & $acc_s$ & $H$ & $acc_u$ & $acc_s$ & $H$ & $acc_u$ & $acc_s$ & $H$ & $acc_u$ & $acc_s$ & $H$\\
\hline
DAP\cite{Lampert2009LearningTD} & 4.2 & 25.1 & 7.2 & 1.7 & 67.9 & 3.3 & 0.0 & 84.7 & 0.0 & 4.8 & 78.3 & 9.0 \\
IAP\cite{Lampert2009LearningTD} & 1.0 & 37.8 & 1.8 & 0.2 & 72.8 & 0.4 & 0.9 & 87.6 & 1.8 & 5.7 & 65.6 & 10.4 \\
CONSE\cite{Norouzi2013ZeroShotLB} & 6.8 & 39.9 & 11.6 & 1.6 & 72.2 & 3.1 & 0.5 & \textbf{90.6} & 1.0 & 0.0 & \textbf{91.2} & 0.0 \\
CMT\cite{Socher2013ZeroShotLT} & 8.1 & 21.8 & 11.8 & 7.2 & 49.8 & 12.6 & 0.5 & 90.0 & 1.0 & 1.4 & 85.2 & 2.8 \\
CMT*\cite{Socher2013ZeroShotLT} & 8.7 & 28.0 & 13.3 & 4.7 & 60.1 & 8.7 & 8.7 & 89.0 & 15.9 & 10.9 & 74.2 & 19.0\\
SSE\cite{Zhang2015ZeroShotLV} & 2.1 & 36.4 & 4.0 & 8.5 & 46.9 & 14.4 & 8.1 & 82.5 & 14.8 & 0.3 & 78.9 & 0.4 \\
LATEM\cite{Xian2016LatentEF} & 14.7 & 28.8 & 19.5 & 15.2 & 57.3 & 24.0 & 11.5 & 77.3 & 20.0 & 0.1 & 73.0 & 0.2 \\
ALE\cite{Akata2013LabelEmbeddingFA} & 21.8 & 33.1 & 26.3 & 23.7 & 62.8 & 34.4 & 14.0 & 81.8 & 23.9 & 4.6 & 73.7 & 8.7 \\
DEVISE\cite{Frome2013DeViSEAD} & 16.9 & 27.4 & 20.9 & 23.8 & 53.0 & 32.8 & 17.1 & 74.7 & 27.8 & 4.9 & 76.9 & 9.2 \\
SJE\cite{Akata2015EvaluationOO} & 14.7 & 30.5 & 19.8 & 23.5 & 59.2 & 33.6 & 8.0 & 73.9 & 14.4 & 3.7 & 55.7 & 6.9 \\
ESZSL\cite{RomeraParedes2015AnES} & 11.0 & 27.9 & 15.8 & 12.6 & 63.8 & 21.0 & 5.9 & 77.8 & 11.0 & 2.4 & 70.1 & 4.6 \\
SYNC\cite{Changpinyo2016SynthesizedCF} & 7.9 & \textbf{43.3} & 13.4 & 11.5 & 70.9 & 19.8 & 10.0 & 90.5 & 18.0 & 7.4 & 66.3 & 13.3 \\
SAE\cite{Kodirov2017SemanticAF} & 8.8 & 18.0 & 11.8 & 7.8 & 54.0 & 13.6 & 1.1 & 82.2 & 2.2 & 0.4 & 80.9 & 0.9 \\
GFZSL\cite{Verma2017ASE} & 0.0 & 39.6 & 0.0 & 0.0 & 45.7 & 0.0 & 2.5 & 80.1 & 4.8 & 0.0 & 83.3 & 0.0 \\
\hline
SP-AEN \cite{Chen18} & 24.9 & 38.6 & 30.3 & 34.7 & 70.6 & 46.6 & 23.3 & 90.9 & 37.1 & 13.7 & 63.4 & 22.6 \\
PSR\cite{Annadani2018PreservingSR} & 20.8 & 37.2 & 26.7 & 24.6 & 54.3 & 33.9 & 20.7 & 73.8 & 32.3 & 13.5 & 51.4 & 21.4 \\
AREN \cite{Xie2019AttentiveRE} & 9.00 & 38.8 & 25.5 & 38.9 & \textbf{78.7} & \textbf{52.1} & 5.6 & 92.9 & 26.7 & 9.2 & 76.9 & 16.4 \\
%GZSRVSE \cite{Zhu2018GeneralizedZR} & - & - & - & 33.4 & 87.5 & 48.4 & 41.6 & 91.3 & 57.2 & 24.5 & 72.0 & 36.6 \\
%f-VAEGAN-D2 \cite{Xian2019fVAEGAND2AF} & 50.1 & 37.8 & 43.1 & 63.2 & 75.6 & 68.9 & 57.1 & 76.1 & 65.2 & - & - & -\\
\hline
IGSC(linear) & 19.1 & 24.6 & 21.5 & 26.5 & 54.2 & 35.6 & 16.5 & 67.4 & 26.4 & 8.4 & 65.4 & 14.9 \\
%Ours-linear(Res152) & 56.7 & 53.0 & 50.9 & 37.8 & 20.2 & 26.7 & 23.0 & 28.6 & 56.2 & 37.9 & 19.9 &     60.9 & 30.0 & 10.8 & 64.5 & 18.4\\
IGSC(nonlinear) & 22.5 & 36.1 & 27.7 & 27.8 & 66.8 & 39.3 & 19.8 & 84.9 & 32.1 & 13.4 & 69.5 & 22.5 \\
IGSC+CS & \textbf{39.4} & 31.3 & \textbf{34.9} & \textbf{40.8} & 60.2 & 48.7 & \textbf{25.7} & 83.6 & \textbf{39.3} & \textbf{23.1} & 58.9 & \textbf{33.2} \\ 
%Ours-nonlinear(Res152) & 59.8 & \textbf{57.3} & 63.3 & 33.8 & \textbf{23.7} & 36.1 & \textbf{28.6} & \textbf{28.6} & \textbf{68.0} & \textbf{40.3} & \textbf{22.7} & 83.9 & \textbf{35.7} & \textbf{14.9} & 67.6 & \textbf{24.4}\\
\end{tabular}
%}
\end{table*}

% table 3
\begin{table*}[ht]
\centering
\caption{Ablation study on effects of different visual models.}\smallskip
\label{tab:ablation}
\begin{tabular}{c|*{12}{c}}
 & \multicolumn{3}{c}{SUN} & \multicolumn{3}{c}{CUB} & \multicolumn{3}{c}{AWA2} & \multicolumn{3}{c}{aPY}\\
Method & $acc_u$ & $acc_s$ & $H$ & $acc_u$ & $acc_s$ & $H$ & $acc_u$ & $acc_s$ & $H$ & $acc_u$ & $acc_s$ & $H$\\
\hline
Res-101 & 22.5 & 36.1 & 27.7 & 27.8 & 66.8 & 39.3 & 19.8 & 84.9 & 32.1 & 13.4 & 69.5 & 22.5\\
Res-152 & 23.7 & 36.1 & 28.6 & 28.6 & 68.0 & 40.3 & 22.7 & 83.9 & 35.7 & 14.9 & 67.6 & 24.4\\
\hline
Res-101+CS & 39.4 & 31.3 & 34.9 & 40.8 & 60.2 & 48.7 & 25.7 & 83.6 & 39.3 & 23.1 & 58.9 & 33.2\\
Res-152+CS & 40.8 & 31.2 & 35.3 & 42.9 & 61.0 & 50.4 & 27.1 & 83.0 & 40.9 & 26.4 & 53.3 & 35.3\\
\end{tabular}
\end{table*}

\begin{comment}  % added on 07/23
\begin{table*}[h]
\centering
\caption{$N_1$ skewness on SUN benchmark.}\smallskip
\label{tab:skewness}
\begin{tabular}{|c|c|c|c|c|}
\hline
\multirow{}{}{} & \multicolumn{2}{c|}{ZSL} & \multicolumn{2}{c|}{GZSL} \\
\cline{2-5} & test (seen)  & test (unseen) & test (seen) & test (unseen) \\
\hline
DeViSE~\cite{Frome2013DeViSEAD} & 2.163 & 2.360 & 2.355 & 2.849\\ 
\hline
IGSC (linear) & 1.073 & 0.685 & 1.219 & \textbf{1.827}\\
\hline
IGSC (nonlinear) & \textbf{1.046} & \textbf{0.380} & \textbf{0.111} & 2.452\\
\hline
\end{tabular}
\end{table*}
\end{comment}

\subsection{Results}

We compared the IGSC method with a variety of standard and generalized ZSL methods as reported in \cite{Xian19TPAMI}. These methods can be categorized into 1) attribute-based: DAP \cite{Lampert2009LearningTD}, IAP \cite{Lampert2009LearningTD}, CONSE \cite{Norouzi2013ZeroShotLB}, SSE \cite{Zhang2015ZeroShotLV}, SYNC \cite{Changpinyo2016SynthesizedCF}; and 2) embedding-based: CMT/CMT* \cite{Socher2013ZeroShotLT}, LATEM\cite{Xian2016LatentEF}, ALE\cite{Akata2013LabelEmbeddingFA}, DeViSE\cite{Frome2013DeViSEAD}, SJE\cite{Akata2015EvaluationOO}, ESZSL\cite{RomeraParedes2015AnES}, SAE\cite{Kodirov2017SemanticAF}, GFZSL\cite{Verma2017ASE}. Performances of the methods are directly reported from the paper \cite{Xian19TPAMI}. The methods under comparison are inductive to both unseen images and unseen semantic vectors. 

%We reported a few variants of the proposed IaSC model to examine: 1) the choice of the semantic classifier (linear and nonlinear), denoted by IaSC (linear) and IaSC (nonlinear) respectively; and 2) the effect of the calibrated stacking rule (with or w/o calibration).
We conducted the ablation study of the proposed IGSC method to examine the forms of label classifier and the effect of calibrated stacking (Equ.~\ref{equ:calibration}): 1) \textbf{IGSC (linear)} uses linear semantic classification (Equ.~\ref{equ:linear}); 2) \textbf{IGSC (nonlinear)} uses nonlinear semantic classification (Equ.~\ref{equ:nonlinear}); 3) \textbf{IGSC+CS} is the full model that uses nonlinear semantic classification and calibrated stacking.

Table~\ref{tab:zsl} shows the conventional ZSL results. The nonlinear variant of IGSC has a superior performance to those of other methods on the CUB dataset and achieves comparable performances on the other datasets. Although GFZSL \cite{Verma2017ASE} has the best performances on SUN and AWA2, this method performs poorly under the GZSL setting. We observe that using a linear label classifier has a slight improvement against the nonlinear one on aPY, which is considered a small-scale dataset in ZSL benchmarks. The two-layer neural network, in general, outperforms a single linear mapping when a large dataset is used. 

Table~\ref{tab:gzsl} shows the generalized ZSL results. In this experiment, recent methods~\cite{Chen18,Annadani2018PreservingSR,Xie2019AttentiveRE} are included for comparison.
The semantics-preserving adversarial embedding network (SP-AEN) \cite{Chen18} is a GAN-based method, which uses an adversarial objective to reconstruct images from semantic embeddings. The preserving semantic relations (PSR) method \cite{Annadani2018PreservingSR} is an embedding-based approach utilizing the structure of the attribute space using a set of relations. Finally, the attentive region embedding network (AREN) \cite{Xie2019AttentiveRE} uses an attention mechanism to construct the embeddings from the part level (i.e., local regions), which consists of two embedding streams to extract image regions for semantic transfer.

By examining the harmonic mean values, the proposed IGSC method consistently outperforms the other competitive methods on three out of the four datasets. We believe the performance gain is achieved because of the novel modeling of image-guided semantic classifiers. This classifier learning paradigm not only has more training pairs (in the scale of the image set) but also allows different ways to separate classes based on the content of the input image. In comparison with attribute based methods which take a two-step pipeline to detect attributes from one image and aggregate the detection results for label prediction, the proposed method optimizes the steps in a unified process. In comparison with recent state-of-the-art methods~\cite{Chen18,Annadani2018PreservingSR,Xie2019AttentiveRE}, the IGSC method is much simpler. The proposed method does not involve advanced techniques such as GAN and attention models, yet it achieves comparable (or superior) performance to those sophisticated methods.

\begin{figure*}[t]
\begin{center}
 \includegraphics[width=0.70\linewidth]{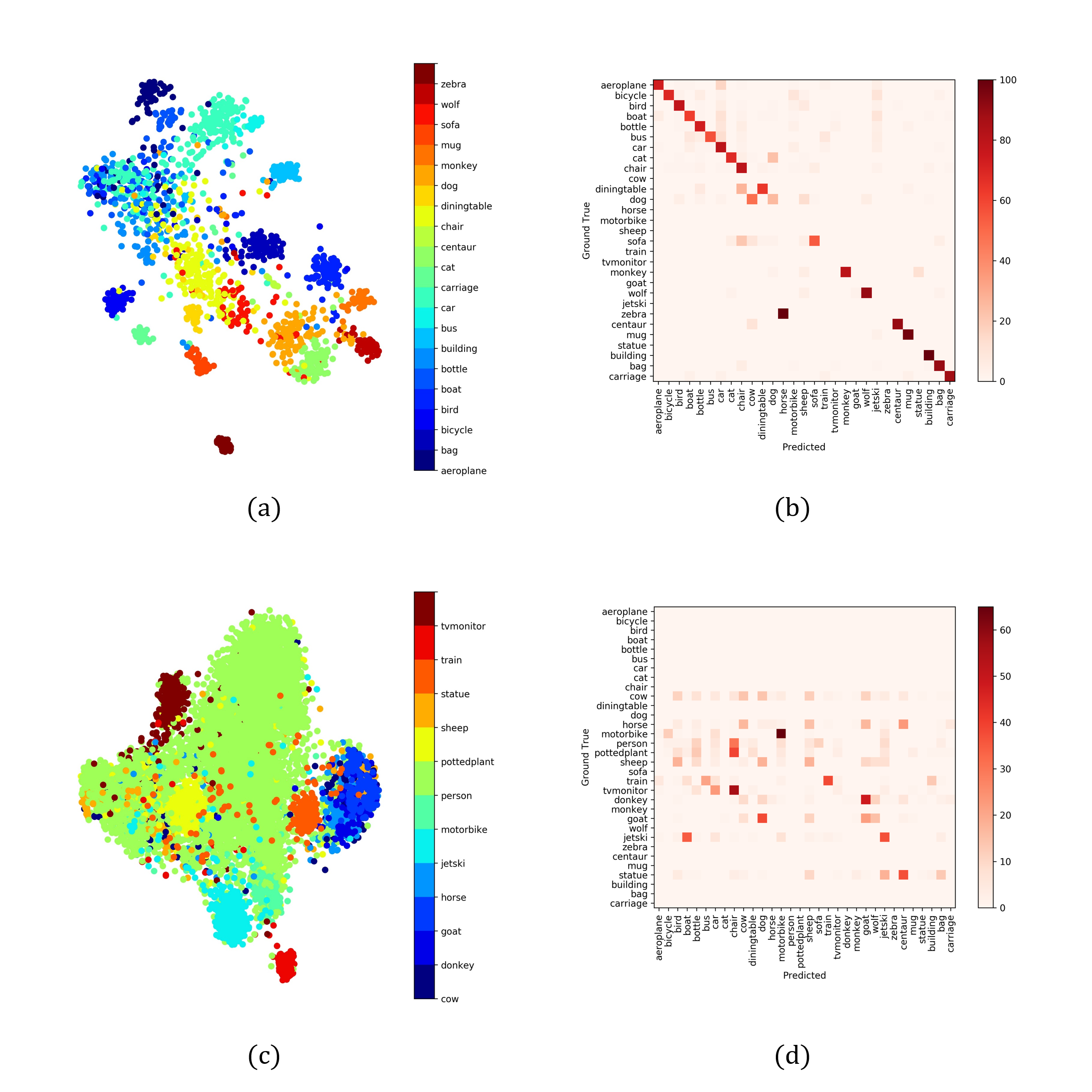}
\end{center}
   \caption{t-SNE visualization of the model space and the confusion matrices in the GZSL experiment using aPY: (top row) seen classes, (bottom row) unseen classes. We decompose the full confusion matrix into two submatrices (seen and unseen classes). Therefore, many elements of the matrices are zeros. Best viewed in color.} 
\label{fig:visualization}
\end{figure*}

\subsection{Analysis and Discussion}

%In this subsection, we discuss the model flexibility and investigate whether hubness can be reduced in IGSC. 
In this subsection, we discuss the model flexibility and visualize the classifier weights generated by IGSC. 

\paragraph{Model extendibility.}
The proposed IGSC model is flexible in that the visual and semantic embeddings, the $h(\cdot)$ and $g(\cdot)$ functions can all be customized to meet specific needs. We provide a proof of concept analysis, in which we investigate the effect of replacing Res-101 with Res-152. Table \ref{tab:ablation} shows the result. Performance improvements are observed when we use a deeper visual model. By elaborating other components in IGSC, it seems reasonable to expect this approach should yield even better results.

\subsection{Visualizing the Label Classifiers}

We visualize the ``model representations'' of the label classifiers by using t-SNE~\cite{t-SNE} of the dynamically generated classifier weights. Figure~\ref{fig:visualization} displays the visualization results and the confusion matrices in the GZSL experiment using the aPY dataset (top row: seen classes, bottom row: unseen classes). Each point in the visualization result represents a label classifier (i.e., a semantic classification model), generated on-the-fly by feeding a test image to the IGSC model. Colors indicate class labels.

Although each image has its own label classifier, the IGSC method tends to generate similar model representations for images of the same class (Fig.~\ref{fig:visualization} (a)). We also observe that a class's intra-class variation of the model representations is related to the prediction performance of that class. For example, the unseen class ``train'' has relatively compact model representations (red points in Fig.~\ref{fig:visualization} (c)). The prediction performance of this class is better than many other classes (Fig.~\ref{fig:visualization} (d)). 

Figure~\ref{fig:visualization} (b) shows an interesting failure case of the model, where images of the seen class ``zebra'' are incorrectly recognized as ``horse'' (an unseen class). One possible reason is the learned ``zebra'' classifiers from seen classes rely significantly on two attributes: rein and saddle, which are not effective attributes for differentiating ``zebra'' and ``horse.'' Indeed, the ``zebra'' model representations (dark red points in Fig.~\ref{fig:visualization} (a)) are gathered together, yet they are quite different from those of the other classes.  

\section{Conclusion}
\label{sec:conclusion}

We propose the IGSC method, which can be used to transform an image into a label classifier, consequently used to predict the correct label in the semantic space. Modeling the correspondence between an image and its label classifier enables a powerful GZSL method that achieves promising performances on four benchmark datasets.

One future research direction we are pursuing is to extend the method for multi-label zero-shot learning, in which images are assigned with multiple labels from an open vocabulary. This would take full advantage of the semantic space. Another direction is to explore model learning with a less restricted setting, which can be transductive for specific unseen classes or test instances.

\clearpage
% ---- Bibliography ----
%
% BibTeX users should specify bibliography style 'splncs04'.
% References will then be sorted and formatted in the correct style.
%
\bibliographystyle{splncs04}
\bibliography{egbib}
\end{document}